\documentclass[10pt, a4paper]{article}

\usepackage{lreccoling2024} 

\usepackage{xurl}
\usepackage{breakurl}
\usepackage{multirow}
\usepackage{multicol}
\usepackage{amssymb}

\def\MLdel#1{\bgroup\markoverwith{\textcolor{brown}{\rule[0.5ex]{2pt}{1pt}}}\ULon{#1}}

\def\MLdel#1{\bgroup\markoverwith{\textcolor{purple}{\rule[0.5ex]{2pt}{1pt}}}\ULon{#1}}

\title{Reproducing the Metric-Based Evaluation of\\a Set of Controllable Text Generation Techniques
}

\name{Michela Lorandi, Anya Belz} 

\address{ADAPT Centre, Dublin City University, Ireland\\
         \{michela.lorandi, anya.belz\}@adaptcentre.ie}

\abstract{
Rerunning a metric-based evaluation should be more straightforward, and results should be closer, than in a human-based evaluation, especially where code and model checkpoints are made available by the original authors. As this
report of our efforts to rerun a metric-based evaluation of a set of single-attribute and multiple-attribute controllable text generation (CTG) techniques shows however, such reruns of evaluations do not always produce results that are the same as the original results,  
and can reveal errors in the reporting of the original work.
 \\ \newline  }

\begin{document}

\maketitleabstract

\section{Introduction} \label{sec:intro}

Over the past few years, the fields of natural language processing (NLP) and machine learning (ML) have seen an increase in interest in reproducibility \cite{sinha2020neurips,branco2020shared,belz2021systematic,belz20232023}. Initially, efforts focussed on promoting and encouraging sharing of all resources needed to rerun experiments, but increasingly it became clear that exact reproduction of results is rarely the outcome even where metric evaluation is concerned. The question is what can be concluded in such situations beyond binary reproduced vs.\ not reproduced findings.

Belz et al.\ (\citeyear{belz2022metrological}; \citeyear{belz20232023}) proposed QRA++, an approach to measuring how close results from two evaluations are, and how reproducible evaluation measures are, in order to facilitate comparison in terms of degree of reproducibility between different methods of evaluation. This approach enables comparable, quantified reproducibility results to be produced.

In this short report, we present our work rerunning the metric-based evaluation of a set of single and multiple-attribute controllable text generation techniques \cite{gu-etal-2022-distributional,gu-etal-2023-controllable}. In the case of all except one pair of scores from the original and reproduction evaluations, the two scores are not the same, and we apply QRA++ to quantify the differences. 

We start with a summary of the QRA++ measures we use (Section~\ref{sec:qra++}), followed by a description of the specific original experiments we repeated in this reproduction study (Section~\ref{sec:orig-work}). We then describe how we went about repeating the work (Section~\ref{sec:repetition}), before presenting the side-by-side results from the original work and our reproduction along with the QRA++ measures of their similarity (Section~\ref{sec:side-by-side}). We finish with some discussion and conclusions (Section~\ref{sec:disc-concl}).

\section{QRA++ Measures}\label{sec:qra++}

QRA++ distinguishes four types of results commonly reported in NLP and ML papers:

\begin{enumerate}\itemsep=0pt
    \item Type I results: single numerical scores, e.g.\ mean quality rating, error count, etc.
\item Type II results: sets of related numerical scores, e.g.\ set of Type I results . 
\item Type III results: categorical labels attached to text spans of any length.
\item Type IV results: Qualitative findings stated explicitly or implied by quantitative results in the original paper.
\end{enumerate}

\noindent The above are quantitatively assessed as follows:

    \begin{enumerate}\itemsep=0pt
    \item Type I results: Small-sample coefficient of variation CV* \cite{belz2022metrological}.
    \item Type II results: Pearson’s r, Spearman’s $\rho$.
    \item Type III results: Multi-rater: Fleiss’s $\kappa$; Multi-rater, multi-label: Krippendorff’s $\alpha$.
    \item Type IV results: Proportion of findings that are / are not confirmed by the repeat experiment. To obtain comparable results we restrict ourselves to pairwise system ranks as findings.
    \end{enumerate}

\noindent In the work reported in this paper we have Type~I, II and IV results, and therefore apply the corresponding quantitative measures above.

\section{Original Work Being Repeated}\label{sec:orig-work}

In the present reproduction study, we carried out repeat evaluations of the main new systems presented by \citet{gu-etal-2022-distributional} and \citet{gu-etal-2023-controllable}. The authors provide the code on GitHub\footnote{\url{https://github.com/HappyGu0524/MultiControl}} and the model checkpoints on Google Drive.\footnote{\url{https://drive.google.com/drive/folders/14XHSG4IAGlAL9t-SYoTUKnAs5ARqHd5f}}

More precisely, the experimental grid we reproduced looks as follows:
\{PriorCTG\} x \{Topic (World, Sports, Business, and Technology), Sentiment (Positive and Negative), Toxicity (Toxic and Non-Toxic)\} x \{no extension, extension\} + \{PriorCTG\} x \{Multi attribute (Topic, Sentiment and Non-Toxic)\} x \{no optim, optim\} + \{MultiCTG\} x \{Multi attribute (Topic, Sentiment and Non-Toxic)\}. The individual systems (MultiCTG, PriorCTG +/- extend/optim) are described in the next section.

\subsection{Systems included in reproduction}

We included the results for the four main new systems from the original work \cite{gu-etal-2022-distributional,gu-etal-2023-controllable} in our reproduction study; we abbreviate system names as follows: MultiCTG, PriorCTG, PriorCTG+extend, and PriorCTG+optim.

\textit{MultiCTG}: This is the core new CTG approach proposed by \citet{gu-etal-2022-distributional} which directly searches for the intersection areas of multiple attribute distributions to achieve control over multiple control attributes. The attribute space is first estimated with an autoencoder structure, then the intersections are iteratively approached via joint minimisation of distances to points representing the controlled attributes. 

\textit{PriorCTG}: This is the core new CTG approach proposed by \citet{gu-etal-2023-controllable}, which utilises a form of latent-space control, more specifically an invertible transformation function,
the Normalizing Flow, that maps the complex
distributions in latent space to simple Gaussian distributions in \textbf{prior} space. 

\textit{PriorCTG+extend}: The \textbf{extend} control strategy additionally achieves opposite control, as in contrastive learning, by using negative weights when interpolating. 

\textit{PriorCTG+optim}: The \textbf{optim} control strategy additionally optimises the intersection of the single attribute representations in prior space to achieve multiple-attribute control.


All systems are trained on the IMDb movie reviews dataset \cite{maas-etal-2011-learning}, the AGNews dataset \cite{zhang2015character}, and the Jigsaw Toxic Comment Classification Challenge Dataset \cite{jigsaw-toxic-comment-classification-challenge}, respectively, for control of sentiment, topic and detoxification attributes. Note that we did not include any of the baseline systems in the reproduction.

\begin{table*}[t!]
    \centering
    \small
    \setlength\tabcolsep{3pt} 
    \renewcommand{\arraystretch}{1.10}
    \begin{tabular}{l|ccc|ccccc|c||c|c}
        \multirow{2}{*}{\textbf{Methods}} & \multicolumn{3}{c|}{\textbf{Sentiment}$\uparrow$ (\%)} & \multicolumn{5}{c|}{\textbf{Topic}$\uparrow$ (\%)} & \textbf{Detox.}$\uparrow$ & \multirow{2}{*}{\textbf{PPL.}$\downarrow$} & \multirow{2}{*}{\textbf{Dist.-1/2/3}$\uparrow$ (\%)} \\
         & \textbf{Avg.} & \textbf{Pos.} & \textbf{Neg.} & \textbf{Avg.} & \textbf{W.} & \textbf{S.} & \textbf{B.} & \textbf{T.} & (\%) &  &  \\
        \hline
        PriorCTG & 97.1 & 99.9 & 94.3 & 95.9 & 95.5 & 99.3 & 90.2 & 98.7 & 90.7 & 61 & 42.0 / 79.7 / 88.4 \\
        PriorCTG Repro & 98.2 & 99.9 & 96.6 & 94.8 & 93.4 & 97.8 & 88.5 & 99.5 & 96.9
        & 59.7 & 41.9 / 79.5 / 88.4 \\
        \hline
        PriorCTG+extend & 99.7 & 99.9 & 99.5 & 97.8 & 97.9 & 99.4 & 94.0 & 99.8 & 95.7 & 61.6 & 42.4 / 79.4 / 88.1 \\
        PriorCTG+extend Repro & 99.3 & 99.9 & 98.7 & 98.2 & 98.2 & 99.5 & 95.5 & 99.8 & 99.9  
        & 60.8 & 42.3 / 79.2 / 88.1 \\ 
    \end{tabular}
    \caption{Side-by-side metric results from original work \cite{gu-etal-2023-controllable} and reproduction study for \textbf{single-attribute control} (last two rows in Table~1 in the original paper). The results of the last two columns are obtained using our own implementation. For PriorCTG and PriorCTG+extend systems (see Section~\ref{sec:orig-work}).  Repro=Reproduction results.}
    \label{tab:res_single}
\end{table*}

\begin{table*}[t!]
    \centering
    \small
    \setlength\tabcolsep{3pt} 
    \renewcommand{\arraystretch}{1.10}
    \begin{tabular}{l|c|ccc||c|c}
        \textbf{Methods} & \textbf{Average}$\uparrow$ (\%) & \textbf{Sentiment}$\uparrow$ (\%) & \textbf{Topic}$\uparrow$ (\%) & \textbf{Detoxification}$\uparrow$ (\%) & \textbf{PPL.}$\downarrow$ & \textbf{Dist.}$\uparrow$ (\%) \\
         \hline
        MultiCTG & 87.4 $\pm$ 10.9 & 86.7 $\pm$ 10.5 & 84.8 $\pm$ 14.2 & 90.7 $\pm$ 7.4 & 31.3 & 59.0 \\
        MultiCTG Repro & 88.4 $\pm$ 8.3 
        & 84.9 $\pm$ 11.5 & 84.5 $\pm$ 14.4 & 95.9 $\pm$ 5.5 
        & 31.5 & 59.2 \\ 
        \hline
        PriorCTG & 89.9 $\pm$ 8.7 & 88.0 $\pm$ 10.6 & 87.4 $\pm$ 8.5 & 94.3 $\pm$ 3.2 & 38.9 & 65.3 \\
        PriorCTG Repro & 91.1 $\pm$ 6.7 
        & 88.0 $\pm$ 10.2 & 87.1 $\pm$ 11.2 & 98.3 $\pm$ 1.6 
        & 38.3 & 65.2 \\
         \hline
        PriorCTG+optim & 92.2 $\pm$ 8.6 & 92.5 $\pm$ 8.5 & 89.3 $\pm$ 11.0 & 94.9 $\pm$ 3.4 & 33.0 & 61.7 \\
        PriorCTG+optim Repro & 93.2 $\pm$ 7.2 
        & 91.8 $\pm$ 9.7 & 89.3 $\pm$ 12.4 & 98.6 $\pm$ 1.1 
        & 32.5 & 62 \\ 
    \end{tabular}
    \caption{Side-by-side metric results from original work \cite{gu-etal-2022-distributional,gu-etal-2023-controllable} and reproduction study for \textbf{multiple-attribute control}. Results for MultiCTG are from the third to last row in \citet{gu-etal-2022-distributional}. Original results for the other two systems are from the last two rows in Table~3 in \citet{gu-etal-2023-controllable}. The results of the last two columns are obtained using our own implementation. For system and metrics descriptions see Section~\ref{sec:orig-work}). Repro=Reproduction results.}
    \label{tab:res_multi}
\end{table*}

\subsection{Evaluation metrics}\label{ssec:eval-metrics}

The metrics in this section are all described in detail in \citet{gu-etal-2022-distributional}. The main set of metrics assesses {single-attribute control performance} (called `attribute relevance' in the original papers), computed as the percentage of outputs that are classified as having the given intended control attribute value by a specific classifier.

For \textit{Sentiment} control performance, the classifier is DeBERTa \cite{he2020deberta} finetuned on the Yelp dataset \cite{zhang2015character}.

For \textit{Topic} control performance, the classifier is DeBERTa finetuned on the AGNews dataset \cite{zhang2015character} utilizing the portion of dataset not used during the model's training. 

For \textit{Toxicity} control performance, there is a discrepancy between what the paper says and what is in the evaluation script shared on GitHub. According to the former, toxicity is measured with the Google Perspective API.\footnote{\url{https://www.perspectiveapi.com/}} However, the script uses a toxicity classifier obtained by finetuning DeBERTa on the Jigsaw Toxic Comment Classification Challenge Dataset,\footnote{\url{https://www.kaggle.com/c/jigsaw-toxic-comment-classification-challenge/}} analogous to control performance assessment for the other control attributes. We ran the evaluation both with Perspective and with the DeBERTa classifier, and found that scores obtained with the latter were closer to the original scores, so those are what we used.

\textit{Multiple-attribute control performance} is computed as the average of the single-attribute control performance scores for the three attributes being controlled.

\textit{Perplexity} is calculated by GPT2-large following the Contrastive Prefix method \cite{qian-etal-2022-controllable}. Note that we used our own implementation as no code was shared for this.

\textit{Distinctness} \cite{li-etal-2016-diversity} is computed as the percentage of distinct n-grams in the continuations generated from a given set of prefixes. System-level 1-gram, 2-gram, and 3-gram distinctness scores are obtained by averaging over prefix-level distinctness scores. In multi-control setting, the average of system-level Distinct-1, 2 and 3 is computed. Here too we used our own implementation based on \citet{yu-etal-2021-attribute-alignment} implementation, because the code was not shared  either by Li et al.\ or by Gu et al.

This gives us six main types of metrics (the three classifier-based metrics, their average (for multiple-attribute control), perplexity, and distinctness). In Table~\ref{tab:res_single} we additionally give the average over the individual control performance scores (\textbf{Avg.} columns) for sentiment, topic and toxicity.

\section{Reproduction Work}\label{sec:repetition}

Our first step was to download the code and model checkpoints from the authors' Github and Drive repositories, and recreate the environments on our machine with a GPU RTXA6000 with 48GB RAM.

We then re-executed the inference phase of the experiments involving PriorCTG from \citet{gu-etal-2023-controllable}, first those with single-attribute control, i.e.\ where Topic, Sentiment or Toxicity are being controlled individually, and then those with multiple-attribute control, where Topic, Sentiment and Toxicity are being controlled at the same time. 

For multiple-attribute control we also re-executed the inference phase of the experiments involving MultiCTG from \citet{gu-etal-2022-distributional}.  This gave us sets of 35 $\times$ 5 $=$ 175 outputs (35 inputs from the PPLM Prompts test set $\times$ 5 repetitions of prompting and collecting the outputs) for each system/attribute combination. 

Note that as in the original work, outputs are generated for all values of all controlled attributes (single-attribute case) or for all combinations of controlled attribute values (multiple-attribute case), results for all of which except Toxicity$=$toxic (`Detox(ification)' in the tables) are reported in the results tables.  In the multiple-attribute case, the average over different attribute value combinations, along with the corresponding standard deviation, is reported.

For the evaluation, we computed the metrics listed in Section~\ref{sec:orig-work}. Recall from Section~\ref{ssec:eval-metrics} that we used the script provided by the authors for Sentiment, Topic and Toxicity control performance assessment. However, we coded our own scripts to compute Perplexity and Distinct-n, as scripts are not provided for these. We also use our own code for the standard deviations in the multiple-attribute table. For all scripts we use parameters as provided by the authors.


Note that as a result of some of the evaluation scripts not being shared, we have two distint reproduction situations (which in QRA++ is reflected in the measurement conditions): (a) for the classifier-based control-performance measures, we use our outputs (regenerated by us using the original authors' code) and evaluate them with the original authors' scripts; and (b) for perplexity and distinctness, we use our outputs \textit{and} our evaluation scripts. In the former case differences in scores can only be due to differences in \textit{executing} the original authors' code, whereas in the latter case, differences can be due to both execution and differences in the evaluation code.

In order to avoid this dual possible source of difference for perplexity and distinctness scores, we decided to re-evaluate the original authors' outputs with our own script. This means that the scores in our tables are not the same as in the two original papers for these two metrics. But it means CV* scores and other reproducibility measures are comparable across all metrics.

\section{Side-by-Side Results and QRA++ Assessment}\label{sec:side-by-side}


Tables~\ref{tab:res_single} and~\ref{tab:res_multi} present side-by-side evaluation results for the original and reproduction work, for each of the six metrics from Section~\ref{sec:orig-work}, plus, in  Table~\ref{tab:res_single} only, averages over individual control performance scores (\textbf{Avg.} columns). Reall that we reevaluated the original authors' outputs in terms of Perplexity and Distinctness (see preceding section).

\begin{table*}
    \setlength\tabcolsep{4pt} 
    \renewcommand{\arraystretch}{1.15}
\small
    \begin{tabular}{|l|c|c|c|c|c|c|c|c|c||c|c|c|c|}
    \hline
\multirow{3}{*}{System}& \multicolumn{13}{c|}{\textit{CV* between original and reproduction scores for each evaluation measure}}\\\cline{2-14}
   & Sent & Sent & Sent & Topic & Topic & Topic & Topic & Topic & Detox & PPL & Dist-1 & Dist-2 & Dist-3 \\
 & avg & pos & neg & avg & W & S & B & T &&&&&\\
    \hline
Prior-CTG & 1.12 & 0 & 2.4 & 1.15 & 2.22 & 1.52 & 1.9 & 0.8 & 6.59 &  2.15	& 0.24	& 0.25	& 0 \\
    \hline
Prior-CTG+ext & 0.4 & 0 & 0.8 & 0.41 & 0.31 & 0.1 & 1.58 & 0 & 4.28 &  1.3	& 0.24	& 0.25	& 0  \\
    \hline
Average & 0.76 & 0 & 1.6 & 0.78 & 1.27 & 0.81 & 1.74 & 0.4 & 5.44 &  1.725	& 0.24	& 0.25	& 0  \\    
    \hline
\end{tabular}
\caption{CV* for each pair of original and reproduction metric scores, for the Prior-CTG and Prior-CTG+extend systems, and the average over both systems.}\label{tab:cv*-single}
\end{table*}
\begin{table*}
    \setlength\tabcolsep{15.75pt} 
    \renewcommand{\arraystretch}{1.15}
\small\centering
    \begin{tabular}{|l|c|c|c|c||c|c|}
    \hline
\multirow{2}{*}{System}       & \multicolumn{6}{c|}{\textit{CV* between original and reproduction scores for each evaluation measure}}\\\cline{2-7}
 & Avg	&Sentiment	&Topic	&Detox	&PPL	&Distinct-n \\
    \hline
Multi-CTG & 1.13 & 2.09 & 0.35 & 5.56 &  0.64	& 0.34  \\
    \hline
Prior-CTG & 1.32 & 0.0 & 0.34 & 4.14 &  1.52	& 0.15 \\
    \hline
Prior-CTG+optim & 1.08 & 0.76 & 0.0 & 3.81 &  1.52	& 0.48 \\
    \hline
Average & 1.18 & 0.95 & 0.23 & 4.5 &  1.23	& 0.32  \\
    \hline
\end{tabular}
\caption{CV* for each pair of original and reproduction metric scores, for the Multi-CTG, Prior-CTG and Prior-CTG+optim systems, and the average over all three.}\label{tab:cv*-multi}
\end{table*}
\subsection{Type IV results}

Regarding Type~IV results (findings), here we are assessing relative performance between systems, such that each pairwise ranking counts as one finding. Note that statistical significance was not computed in the original work. 

For single-attribute control (Table~\ref{tab:res_single}), in the original work, Prior~CTG+extend has higher scores than PriorCTG according to all metrics except for Perplexity and 2-gram and 3-gram Distinctness where PriorCTG scores are very slightly higher. For Sentiment/Pos, scores are identical. In our reproduction evaluations, these two systems are ranked the same way in all cases, giving us a perfect proportion of 13/13 findings upheld for this table.

For multiple-attribute control (scores in  Table~\ref{tab:res_multi}), the same type of analysis gives us a proportion of 18/18 findings upheld (pairwise ranks confirmed).

\subsection{Type I results}

For Type~I results, we computed CV* values for all individual system/metric level original and reproduction scores. We report the individual scores, as well as the mean per metric.

For single-attribute control (scores in Table~\ref{tab:res_single}), Table~\ref{tab:cv*-single} shows CV* scores for each pair of original and reproduction metric scores, for the Prior-CTG and Prior-CTG+extend systems, and the average over both systems (last row). 

One clear tendency is that the Prior-CTG system has better reproducibility scores across the board than Prior-CTG+extend (except for distinctness metrics where the two systems are tied).

Looking at metric-level differences (`Average' row), we can see that Perplexity and (by a smaller margin) Detoxification Control  have lower reproducibility than the other metrics.



For multiple-attribute control (scores in Table~\ref{tab:res_multi}), Table~\ref{tab:cv*-multi} shows CV* scores for each pair of original and reproduction metric scores, for the Multi-CTG, Prior-CTG and Prior-CTG+optim systems, and the average over all three (last row). We can see that here too, the Perplexity and Detoxification Control metrics have the poorest reproducibility.

We can also see a slight tendency for the classifier scores for the Prior-CTG+optim system to have better reproducibility than the other two systems (but not for PPL and Distinct-n), but the picture is more mixed than for the single-attribute control systems.

\subsection{Type II results}

For Type II results we compute Pearson's correlation coefficients between sets of metric scores in two ways, (i) for each metric (i.e.\ how do all the scores for each metric correlate between original and reproduction), and (ii) for each system (i.e.\ how do all the scores for each system correlate).

For single-attribute control (scores in Table~\ref{tab:res_single}), system-level Pearson's between all metric results in the original and reproduction runs is above 0.99 for both Prior-CTG and Prior-CTG+extend. Mean metric-level Pearson's is perfect (but note that we have only two score pairs all of which are ranked identically).

For multiple-attribute control (scores in Table~\ref{tab:res_multi}), system-level Pearson's between all metric results in the original and reproduction runs is  above 0.99 for all three systems. Metric-level Pearson's is  above 0.99 for all metrics except the sentiment-classifier metric which at $r= $ 0.969 is slightly lower than the other metrics. Mean metric-level $r$ is 0.994.

\section{Discussion and Conclusion}\label{sec:disc-concl}

The main challenges in carrying out our reproduction study were (i) lack of clarity in the paper with respect to what the averages and standard deviations in results tables were computed over, and (ii) discrepancies between the shared code and what the paper said, e.g.\ the paper says toxicity was assessed with Perspective, whereas the shared evaluation script has a toxicity classifier.  

Our quantified reproducibility assessments revealed a high degree of reproducibility at the study level for Type~II and Type~IV results. For Type~I results, study-level CV* (computed as the mean of metric-level means) was 1.154 for single-attribute control, and 1.402 for multiple-attribute control. While this compares well to reproducibility results in human evaluations which very rarely achieve study-level CV* below 5 in pairwise comparisons of original study and one reproduction, it does confirm once again that even with identical code, we cannot necessarily expect to get the same results.


In terms of metric-level CV*, the Detoxification control  metric had notably worse reproducibility than the others which may be partly but not entirely explainable by the fact that only Toxicity=nontoxic was taken into account here. 



In terms of the results that tend to be considered as most important, Type IV results or findings upheld, reproducibility was perfect with all pairwise rankings being identical in the original and reproduction experiments. 

\nocite{*}
\section*{Bibliographical References}\label{sec:reference}

\bibliographystyle{lreccoling2024natbib}
\bibliography{biblio}

\appendix

\section{Perplexity and Distinct-n implementation}
No code was shared to compute perplexity and Distinct-n, hence we used our own implementation. Perplexity is calculated using the evaluate library of HuggingFace\footnote{\url{https://huggingface.co/docs/evaluate/en/index}} using GPT-2 Large.

System-level Distinct-n (n=1, 2, 3) is the average Distinct-n at prefix-level, which is computed as the number of unique n-grams in the set of generated outputs with the same prefix over the total amount of tokens. GPT-2 is used to tokenise the texts.

Table~\ref{tab:res_single_orig} and~\ref{tab:res_multi_orig} show Perplexity and Distinct-n results reported in the original work, the results of the original study computed using our implementation and the reproduction using our implementation.

\begin{table*}[h!]
    \centering
    \small
    \setlength\tabcolsep{3pt} 
    \renewcommand{\arraystretch}{1.10}
    \begin{tabular}{l|c|c}
        \textbf{Methods} & \textbf{PPL.} (\%) $\downarrow$ & \textbf{Dist.-1/2/3}$\uparrow$ (\%) \\
        \hline
        PriorCTG & 54.3 & 29.1 / 70.1 / 86.9 \\ 
        PriorCTG using our implementation & 61 & 42.0 / 79.7 / 88.4 \\ 
        PriorCTG Repro & 59.7 & 41.9 / 79.5 / 88.4 \\
        \hline
        PriorCTG+extend & 54.6 & 29.8 / 70.5 / 86.8 \\ 
        PriorCTG+extend using our implementation & 61.6 & 42.4 / 79.4 / 88.1 \\ 
        PriorCTG+extend Repro & 60.8 & 42.3 / 79.2 / 88.1 \\ 
    \end{tabular}
    \caption{Side-by-side metric results from original work \cite{gu-etal-2023-controllable}, original work \cite{gu-etal-2023-controllable} computed using our own implementation and reproduction study using our own implementation for \textbf{single-attribute control} (last two rows in Table~1 in the original paper). For PriorCTG and PriorCTG+extend systems (see Section~\ref{sec:orig-work}).  Repro=Reproduction results.}
    \label{tab:res_single_orig}
\end{table*}

\begin{table*}[h!]
    \centering
    \small
    \setlength\tabcolsep{3pt} 
    \renewcommand{\arraystretch}{1.10}
    \begin{tabular}{l|c|c}
        \textbf{Methods} & \textbf{PPL.}$\downarrow$ & \textbf{Dist.}$\uparrow$ (\%) \\
         \hline
        MultiCTG & 28.4 & 49.5 \\ 
        MultiCTG using our implementation & 31.3 & 59.0 \\ 
        MultiCTG Repro & 31.5 & 59.2 \\ 
        \hline
        PriorCTG & 34.7 & 55.5 \\ 
        PriorCTG using our implementation & 38.9 & 65.3 \\ 
        PriorCTG Repro & 38.3 & 65.2 \\
         \hline
        PriorCTG+optim & 29.6 & 51.6 \\
        PriorCTG+optim using our implementation & 33.0 & 61.7 \\
        PriorCTG+optim Repro & 32.5 & 62 \\ 
    \end{tabular}
    \caption{Side-by-side metric results from original work \cite{gu-etal-2022-distributional,gu-etal-2023-controllable}, original work \cite{gu-etal-2022-distributional,gu-etal-2023-controllable} computed using our own implementation and reproduction study using our own implementation for \textbf{multiple-attribute control}. Results for MultiCTG are from the third to last row in \citet{gu-etal-2022-distributional}. Original results for the other two systems are from the last two rows in Table~3 in \citet{gu-etal-2023-controllable}. For system and metrics descriptions see Section~\ref{sec:orig-work}). Repro=Reproduction results.}
    \label{tab:res_multi_orig}
\end{table*}


\end{document}